\documentclass[runningheads]{llncs}
\usepackage[T1]{fontenc}
\usepackage{amsmath}
\usepackage{amssymb}

\usepackage{enumitem}
\usepackage{graphicx}
\usepackage[table]{xcolor}
\usepackage{subcaption}

\def\be{\begin{equation}}
\def\ee{\end{equation}}
\def\bea{\begin{eqnarray}}
\def\eea{\end{eqnarray}}

\begin{document}

\title{Evidential Physics-Informed Neural Networks \\for Scientific Discovery}
\titlerunning{E-PINN for Scientific Discovery}
%
\author{Hai Siong Tan\inst{1} \and
Kuancheng Wang\inst{2} 
\and
Rafe McBeth\inst{3}}
\authorrunning{Tan et al.}
%
\institute{Gryphon Center for A.I. and Theoretical Sciences, Singapore \\
\email{haisiong.tan@gryphonai.com.sg}
\and
Electrical and Computer Engineering Department, \\University of California San Diego, CA, USA
\and
University of Pennsylvania, Perelman School of Medicine,\\
Department of Radiation Oncology, Philadelphia, USA
}
%
\maketitle              
\begin{abstract}
We present the fundamental theory and implementation guidelines underlying Evidential Physics-Informed Neural Network (E-PINN) -- a novel class of uncertainty-aware PINN. It leverages the marginal distribution loss function of evidential deep learning for estimating uncertainty of outputs, and infers unknown parameters of the PDE via a learned posterior distribution. Validating our model on two illustrative case studies -- the 1D Poisson equation with a Gaussian source and the 2D Fisher-KPP equation, we found that E-PINN generated empirical coverage probabilities that were calibrated significantly better than Bayesian PINN and Deep Ensemble methods. To demonstrate real-world applicability, we also present a brief case study on applying E-PINN to analyze clinical glucose-insulin datasets that have featured in medical research on diabetes pathophysiology.

\end{abstract}

\section{Introduction}
\setcounter{footnote}{0}
A central problem in scientific machine learning is the inference of latent and unknown parameters governing the physical system from observed data. This has typically been addressed through a variety of methods, including \emph{Physics-Informed Neural Networks} (PINN) \cite{Lagaris_1998,pinn} when there is a presumed description via differential equations, neural ODE systems \cite{neural} which embeds neural networks within ODE solvers, and Bayesian neural networks that could incorporate known mathematical models (e.g. B-PINN \cite{bpinn}). 

Most recently, Evidential Physics-Informed Neural Networks (E-PINN) was proposed in \cite{tan3},
where principles of \emph{Evidential Deep Learning} (EDL)\footnote{Evidential Deep Learning \cite{amini,sensoy} is a framework where a higher-order set of priors are placed over likelihood functions for the outputs, and model training is then used to infer uncertainty estimates from closed-form functions of the learned prior hyperparameters.} was blended with those of PINN to yield uncertainty estimates for the target outputs assumed to follow a set of PDEs. E-PINN \cite{tan3} was conjectured to be able to generate uncertainty estimates also for the unknown PDE parameters (apart from the target output) by averaging the PDE residual term over a hypothetical distribution. A limitation of the work of \cite{tan3} is that it only applies to a subclass of PDEs where its parameters can be separated out as linear terms in a certain form (as expressed in equation 3 of \cite{tan3}). It was also not clear how one should specify the hypothetical distribution for the learnable PDE parameters.  

In our work here, we present a significant refinement of the framework of \cite{tan3}, in particular, proposing a novel approach for inferring the unknown parameters that is much more aligned with Bayesian statistics.  
We formulate a data-driven, information-theoretic approach to specify a prior distribution for the parameters. Unlike ordinary PINN or the work in \cite{tan3}, the loss weight of the PDE residual is not fixed a priori but is learned adaptively, evolving throughout training. This refinement requires a two-phase model training algorithm where model is first trained purely on data in the first phase, allowing us to construct priors and appropriate initial conditions for the PDE residual and the learnable parameters, before invoking the PDE constraints in the second phase.   

As validation case studies, we applied our model to a 1D Poisson equation with Gaussian source and a 1+1D nonlinear Fisher-KPP equation, a prototype of the reaction-diffusion model. We compared the performance of our model against two popular uncertainty-aware frameworks that could also be synthesized with PINN. 
As reviewed in \cite{cuomo2022}, a state-of-the-art proposal for uncertainty quantification in PINNs is the framework of Bayesian-PINN (B-PINN) of \cite{bpinn}, where
the PDE residual loss forms part of the posterior distribution. Another popular framework is that of \emph{Deep Ensemble} (\cite{Ganaie,lak}) where a set of similar models with different initial weights are taken to provide a distribution of predictions. The main performance metric we adopt here is the degree of calibration of the uncertainty quantification measured by the empirical coverage probability \cite{ECP}. To illustrate its applicability in a real-world setting, we also apply E-PINN on published intravenous glucose tolerance test (IVGTT) data assuming the Bergman minimal model of glucose-insulin dynamics \cite{bergman2,bergman1} which has played a crucial role in modeling diabetes pathophysiology \cite{bergman3}. 

Our paper is organized as follows. We begin with an exposition of our theoretical derivations in Sec.~\ref{sec:theory}. In Sec.~\ref{sec:case}, we
validate our model on two case studies, followed by an application of E-PINN on a real-world clinical dataset in Sec.~\ref{sec:Bergman},
before concluding in Sec.~\ref{sec:conclusion}.

\section{Theoretical Formulation}
\label{sec:theory}
\subsection{A posterior density from merging principles of Deep Evidential Regression and PINN}
Our loss function is the negative logarithm of a probability density-like function
that can be schematically expressed as follows.
\be
\label{loss}
\mathcal{L} = -\log\left[
P(\mathcal{D} | \mathcal{M}(\vec{w}) ) \,
P(\mathcal{M}(\vec{w}) | \vec{\Omega} )  \,
\pi ( \vec{\Omega} ) \right], 
\ee
where $\mathcal{M}$ refers to a neural network providing a surrogate model for the empirical dataset $\mathcal{D} = \{ 
\left( 
\vec{x}_i, \vec{y}_i \right) 
\}_{i=1}^N,
$
with $\vec{w}$ being its weights, and $\vec{\Omega}$
denoting the set of unknown parameters of an underlying PDE description of data $\mathcal{D}$.
In this current work, we take
$\mathcal{M}$
to be a multilayer feedforward neural network. In \eqref{loss}, 
the term $P(\mathcal{M}(\vec{w}) | \vec{\Omega} )$ refers to the exponential of (the negative of) the PDE residual term in PINN. 
\be
\label{resi_exp}
P(\mathcal{M}(\vec{w}) | \vec{\Omega} ) \sim
\exp{ \left[- \frac{1}{2\sigma^2_R}
\sum^{N_D}_{k=1}
\mathcal{R}^2_k \left(
\partial y, y, x_k, \vec{\Omega}
\right) \right]
},
\ee
where $N_D$ is the number of independently sampled points within the domain of the PDE and $
\mathcal{R} \left(
\partial y, y, x, \vec{\Omega}
\right) = 0$
refers to the partial differential equation, with $\sigma^2_R$ being the variance parameter of a density function for the residual.\footnote{
One can extend this to a system of PDEs by defining a separate residual for each differential equation.} The negative logarithm of \eqref{resi_exp}
is the PDE residual loss, with
the parameter $\sigma^2_R$ quantifying the weight of this loss term relative to the data loss term $\sim -\log P(\mathcal{D}| \mathcal{M} (\vec{w}) )$. In standard PINN, this is a free parameter and, to our knowledge, there is no principled approach towards determining its choice. Here we lift $\sigma^2_R$ to be a learnable parameter with the relative weight of the PDE residual evolving as the model shifts towards a minimum in the loss landscape. 
We regularize the dynamical evolution of $\sigma^2_R$ through a prior density function $\pi (\sigma^2_R; \alpha_r, \beta_r)$ which we pick to be the inverse-gamma distribution. Since the residual is a function of the unknown parameters $\vec{\Omega}$, the form of \eqref{loss} suggests that 
we can interpret it as the (unnormalized) posterior density for $\vec{\Omega}$, with 
$\pi (\vec{\Omega} )$ being the prior distribution and the product
$P(\mathcal{D} | \mathcal{M}(\vec{w}) ) \,
P(\mathcal{M}(\vec{w}) | \vec{\Omega} )$ being the likelihood function. The model's weights are then latent variables with model training equivalent to a \emph{maximum a posteriori} estimation. This interpretation enables us to compute the uncertainty of $\vec{\Omega}$ as being defined with respect to the posterior density function
\be
\label{posterior}
f_p \left(\vec{\Omega} | \mathcal{D}, \mathcal{M}(\vec{w}) \right)
= \frac{P(\mathcal{M}(\vec{w})|\vec{\Omega}) \pi(\vec{\Omega})}{\int d\vec{\Omega}\,
P(\mathcal{M}(\vec{w})|\vec{\Omega}) \pi(\vec{\Omega})
},
\ee
where we have used the fact that 
the likelihood function factorizes
\be
\label{factoredPosterior}
P(\mathcal{D}, \vec{w} | 
\vec{\Omega} ) = 
P(\mathcal{D} | \mathcal{M}(\vec{w}) )P(\mathcal{M}(\vec{w}) | \vec{\Omega} ),
\ee
and thus the data loss term does not appear explicitly, being canceled away in the normalization factor. Restoring the input indices, we note that since the data loss term and PDE residual term are products of i.i.d. individual observations, \eqref{factoredPosterior} represents 
\be
P(\mathcal{D} | \mathcal{M}(\vec{w}) )P(\mathcal{M}(\vec{w}) | \vec{\Omega} )
\equiv
\prod^{N_D}_{j=1}
P(\mathcal{D}_j| \mathcal{M}(\vec{w}), x_j )
\prod^{N_p}_{k=1}
P(\mathcal{M}(\vec{w}) | \vec{\Omega}, x_k ),
\ee
where $N_D$ is the total number of empirically observed targets and $N_p$ is the number of points in the domain of the PDEs upon which we chose to condition the model on. In general, the choice of $\{x_k \}_{k=1}^{N_p}$ defines the set of discrete input values where we assert the model to be close to the presumed PDEs.

Now we take $P(\mathcal{D} | \mathcal{M}(\vec{w}) )P(\mathcal{M}(\vec{w}) | \vec{\Omega} )$ as a joint likelihood function for $\vec{\Omega}$ which defines a posterior distribution for $\vec{\Omega}$ following eqn.\eqref{posterior}. 
For the data loss term, we take it to be the term proposed in `Deep Evidential Regression' \cite{amini} which maps a generic input variable to four output functions $\{\gamma, \alpha, \beta, \nu \}$. It is a t-distribution with density function
\be
\label{EDL_density}
P(\mathcal{D} | \mathcal{M}(\vec{w}); 
\gamma, \alpha, \beta, \nu) 
= \frac{\Gamma\left( \alpha + \frac{1}{2} \right)}{\Gamma \left( \alpha  \right)  \sqrt{2\pi \beta  (1+\nu)/\nu  }}
\left(
1 + \frac{(y - \gamma)^2}{2\beta  (1+\nu)/\nu}
\right)^{-(\alpha+\frac{1}{2})},
\ee
where $\gamma$ is the mean 
output and $\alpha, \beta, \nu$ are learnable functions of the input, related to the overall predictive uncertainty $\sigma^2_p$ as follows.  
\be
\label{NIG_defin}
\gamma = \mathbb{E}[Y], \,\,
\sigma^2_p = \text{Var}[Y] 
= \frac{\beta}{\alpha - 1} \left( 1 + \frac{1}{\nu} \right),
\ee
where $Y$ denotes each output variable 
in the dataset $\mathcal{D}$, and 
the expectation and variance are taken with respect to the t-distribution density function in \eqref{EDL_density}. 
This distribution is the marginal distribution obtained after integrating the mean and variance of a Gaussian likelihood function $\mathcal{N}(\mu, \sigma^2)$ with respect to a normal-inverse-gamma prior distribution, where
$\mu \sim \mathcal{N}(\gamma, \sigma^2/\nu)$, $\sigma^2 \sim \Gamma^{-1}(\alpha, \beta)$. 
We take the dependent variable in the PDE residual to be the mean output, i.e. $y \rightarrow \gamma (x)$ in eqn.~\eqref{resi_exp}. This implies that the MAP estimation involves the adherence of the \emph{mean} target variables to the PDEs. Upon completion of model training, we can place confidence intervals on model's predictions using eqn.~\eqref{NIG_defin}. Since we infer $\vec{\Omega}$ at the end of model training via eqn. \eqref{posterior}, our framework thus appears as a \emph{maximum a posteriori} estimation of $\vec{\Omega}$, or more precisely a maximum likelihood estimation that is regularized by the prior $\pi (\vec{\Omega})$. 

\subsection[uncertainty of omega]{On uncertainty of $\vec{\Omega}$ and its prior distribution }

In our framework, we alluded to a
posterior density function
$f_p \left(\vec{\Omega} | \mathcal{D}, \mathcal{M}(\vec{w}) \right)$ in eqn. \eqref{posterior} 
of which negative logarithm is the model's loss function. 
The uncertainty in $\vec{\Omega}$ reflects the probabilistic deviation of the model from the PDE description, as quantified by the residual. In the following, we will invoke this principle to derive a form for the prior $\pi (\vec{\Omega})$. 
Let $\mathcal{D}_{\vec{\Omega}}$ denote the finite, discretized domain for the unknown parameters $\vec{\Omega}$. At each point of $\mathcal{D}_{\vec{\Omega}}$, we can evaluate the mean squared deviation between the numerical solution to the PDE characterized by $\vec{\Omega}$ ($L_p (x; \vec{\Omega})$) and the data-fitted model $\mathcal{M}_0 (x; \vec{w}_0)$. 
\be
\label{msd}
M (\vec{\Omega}) = \frac{1}{N_D}\sum^{N_D}_{j=1}
\left(
L_p (x_j; \vec{\Omega}) - 
\mathcal{M}_0 (x_j; \vec{w}_0)
\right)^2.
\ee
We assert a Gaussian likelihood based on the above mean squared deviation for $\vec{\Omega}$, with the variance parameter being the mean $\overline{M}$ averaged over the domain $\mathcal{D}_{\vec{\Omega}}$. This defines a density function $f$ at each point $\Omega$ of the form 
\be
\label{fund_prior}
f(\vec{\Omega}) = \frac{1}{N} e^{-\frac{M(\vec{\Omega})}{2\overline{M}}},\,\,\, N = \int_{\mathcal{D}_{\vec{\Omega}}}
d\vec{\Omega}\, f(\vec{\Omega}),\,\,\,
\overline{M} \equiv 
\frac{1}{|\mathcal{D}_{\vec{\Omega}}|}
\int_{\mathcal{D}_{\vec{\Omega}}}
d\vec{\Omega}\,
M(\vec{\Omega}).
\ee
where $N$ is the normalization constant and all integrals are  implemented as numerical Riemann sums over the discretized 
domain $\mathcal{D}_{\vec{\Omega}}$. 
We would like the prior distribution of $\Omega$ to be characterized by the same mode and dispersion scales as the highest density region (HDR) of $f$ as discussed in \cite{Hyndman}. At some confidence level, say 68\%, this region is generally a complex subset of the domain $\mathcal{D}_\Omega$. Since our choice of prior distribution affects model training dynamics in the second phase, we adopt a simple Gaussian surrogate distribution for this region, with the means being the modes 
and the standard deviations being those of each marginal distribution. 
\be
\label{prior_formula_1}
\pi (\vec{\Omega}; \vec{\mu}, \Sigma ) 
\sim \frac{1}{\sqrt{\det \Sigma}}
\text{exp}\left[- 
\frac{
\lvert \lvert
\vec{\Omega} - \vec{\mu} \rvert
\rvert^2}
{2\Sigma}
\right],
\ee
where $\Sigma$ is a diagonal covariance matrix of which elements are the variances of the marginal distribution for each component of $\vec{\Omega}$,
while the mean vector $\vec{\mu}$ are the modes of $f(\vec{\Omega})$
\be
\label{stats_prior}
\vec{\mu} = 
\underset{\vec{\Omega}}{\arg \max}\, f(\vec{\Omega}),\,\,\,
\Sigma_{ij} = \delta_{ij} \text{Var} \left[
\int_{\mathcal{D}_{\vec{\Omega}}}
d\Omega_1 \ldots 
d\Omega_{i-1} d\Omega_{i+1}\ldots d\Omega_m
\,\,\,\,\, f(\vec{\Omega}) \right].
\ee
This choice of the prior distribution yields a simple approximation of the highest density region of $f(\vec{\Omega})$ (eqn.\eqref{fund_prior}) which is in turn based on the mean squared deviation between the data-fitted model's curve and the numerical solution equipped with $\vec{\Omega}$, with the dispersion scale in each parameter component $\Omega_k$ set by the variance of its marginal distribution. 

\subsection[det]{Determination of $\pi(\sigma^2_R; \alpha_r, \beta_r)$}
\label{sec:fix_sigmaR}

Having defined the prior density for $\vec{\Omega}$, we now invoke an information-theoretic argument to set the prior density for $\sigma^2_R$ -- the learnable weight for the PDE residual loss term. Its prior density is intended to
guide the evolution of $\sigma^2_R$ during the gradient descent-based training as the model adapts to both data and PDE constraints. Assuming an inverse-gamma distribution for its form, 
\be
\label{prior_R}
\pi(\sigma^2_R; \alpha_r, \beta_r) =
\frac{\beta_r^{\alpha_r}}{\Gamma (\alpha_r)}
\sigma^{-2(\alpha_r + 1)}_R e^{-\frac{\beta_r}{\sigma^2_R}},
\ee
we pick its hyperparameters $(\alpha_r, \beta_r)$ such that it is consistent with other aspects of our formalism. The mode and mean values are
$\frac{\beta_r}{\alpha_r + 1}$ and $\frac{\beta_r}{\alpha_r - 1}$ respectively. Here we restrict ourselves to the case where $\alpha_r >1$ so that the mean is well-defined. 
We pick the initial value of $\sigma^2_R$ ($\equiv \sigma^2_{ini}$) to be the mean of \eqref{prior_R}. As $\sigma^2_R$ decreases during model training, it approaches the mode of $\pi (\sigma^2_R; \alpha_r, \beta_r)$ at which the derivative with respect to $\sigma^2_R$ vanishes. 
\be
\sigma^2_{ini} = \frac{\beta_r}{\alpha_r - 1}, \,\,\,
\sigma^2_{asy} =\frac{\beta_r}{\alpha_r + 1},
\ee
where $\sigma^2_{ini}$ denotes initial value, and $\sigma^2_{asy}$ denotes an asymptotic lower bound at the completion of model training. 
Since the distribution of $\vec{\Omega}$ is defined through eqn.\eqref{posterior}, preceding model training, we would like the initial likelihood function to be close to the prior distribution for $\vec{\Omega}$. This motivates setting $\pi(\sigma^2_R; \alpha_r, \beta_r)$ such that the initial induced statistics of $\vec{\Omega}$ is similar to $\pi (\vec{\Omega}; \vec{\mu}, \Sigma)$. We use the Kullback-Leibler divergence as a measure of similarity, and set
\be
\label{sigmaR} 
\frac{\beta_r}{\alpha_r - 1} = 
\underset{\sigma^2}{\arg \min}\,\,\, D_{KL} \left(
P(\mathcal{M}_0 (\vec{w}^0)| \vec{\Omega} ; \sigma^2) \Vert 
\pi(\vec{\Omega}; \vec{\mu},\Sigma )
\right),
\ee
where 
\be
\label{ini_sigmaR}
P(\mathcal{M}_0 (\vec{w}^0)| \vec{\Omega} ; \sigma^2) = 
\frac{\text{exp} \left[- \frac{1}{2\sigma^2}
\sum^{N_D}_{k=1}
\mathcal{R}^2_k \left(
\partial f, f, x_k, \vec{\Omega}
\right) \right]}{\int d\vec{\Omega}\,\,\,\text{exp} \left[- \frac{1}{2\sigma^2}
\sum^{N_D}_{k=1}
\mathcal{R}^2_k \left(
\partial f, f, x_k, \vec{\Omega}
\right) \right]}.
\ee
More intuitively, the parameter $\sigma^2_R$ controls the overall scale of the dispersion of each component of $\vec{\Omega}$. The constraint \eqref{sigmaR} sets the initial $\sigma^2_R$ such that the likelihood function 
is initially close (in the sense of KL measure) to the prior function for $\vec{\Omega}$. 

As the model adapts to the PDE residual condition, $\sigma^2_R$ decreases and moves from the mean towards the mode where the derivative with respect to $\sigma^2_R$ vanishes. We would like the minimum uncertainties at this point to be consistent with our model implementation, in particular, the discrete nature of the domains for the parameters $\vec{\Omega}$. These domains are necessarily characterized by finite resolutions. Consider a diagonal multivariate Gaussian distribution $\pi_m (\vec{\Omega}; \vec{\mu}, \Sigma_{min})$ where each standard deviation of $\Sigma_{min}$ is set as the minimal spacing in each parameter's domain. This then yields a natural choice for the mode of $\pi(\sigma^2_R; \alpha_r, \beta_r)$.
\be
\label{sigmaR_mode} \text{mode} \left( \sigma^2_R \right)  = \frac{\beta_r}{\alpha_r + 1} = 
\underset{\sigma^2}{\arg \min}\,\,\, D_{KL} \left(
P(\mathcal{M}_0 (\vec{w}^0)| \vec{\Omega} ; \sigma^2) \Vert 
\pi_m(\vec{\Omega}; \vec{\mu}, \Sigma_{min} )
\right).
\ee
The two KL divergence-minimization equations \eqref{sigmaR} and \eqref{sigmaR_mode} then determine $\alpha_r, \beta_r$ which regularizes the adaptive evolution of the PDE residual loss term weight $\sigma^2_R$. 


\subsection{A summary of model implementation }
\label{sec:guide}

Our framework is structured around a two-phase training algorithm where in the first phase, the neural network is trained purely on the empirical dataset. The loss function in this training phase is simply the negative log-likelihood of eqn. \eqref{EDL_density}.
Upon convergence of the data-fitted model, we then construct the multivariate
Gaussian $\pi (\vec{\Omega})$ as defined in eqn.\eqref{prior_formula_1}
which will serve as the prior distribution. This is done by first specifying the parameters' domains and computing the various quantities in eqns.\eqref{fund_prior} and \eqref{stats_prior}. We also solve
for $\alpha_r, \beta_r$ using \eqref{sigmaR}, \eqref{sigmaR_mode}, after computing minima of the KL divergence term on the RHS of eqns. \eqref{sigmaR} and \eqref{sigmaR_mode}.

The second phase of model training now begins having determined the priors for the parameters and $\sigma^2_R$. We now train the model using the full loss function 
\be
\mathcal{L} = -\log\left[
P(\mathcal{D} | \mathcal{M}(\vec{w}) ) \,
P(\mathcal{M}(\vec{w}) | \vec{\Omega} )  \, 
\pi (\sigma^2_R; \alpha_r, \beta_r) \,
\pi ( \vec{\Omega}; \vec{\mu}, \Sigma ) \right], 
\ee
with each factor defined in eqns.\eqref{resi_exp}, \eqref{EDL_density}, \eqref{prior_formula_1} and \eqref{prior_R}. This second phase of training refines the purely data-fitted model such that it conforms to the presumed PDE description. Upon completion, we can obtain confidence bands on the model predictions, and infer $\vec{\Omega}$ with uncertainty as defined by the mean and variance of the posterior \eqref{posterior}. 
Our methodology is broadly applicable to systems of differential equations, irrespective of the choice of differential operators or dimensionality, as reflected in the nature of eqns. \eqref{resi_exp}, \eqref{EDL_density}, \eqref{prior_formula_1} and \eqref{prior_R}.

\section{Validation case studies: 1D Poisson and 2D Fisher-KPP equations }
\label{sec:case}

We validated the E-PINN framework using two representative differential equations : (i)the 1D Poisson equation with Gaussian source (ii)the 
Fisher-KPP equation. These prototype equations and their variants have found wide applicability across scientific disciplines (see e.g. \cite{Poisson1,Fisher1,Fisher2,Poisson2}.) For the main performance metric, we adopt the empirical coverage probability (ECP) to measure how well-calibrated the uncertainties are, relative to the nominal ones (NCP). In our case studies, we computed plots of ECP vs NCP and calculated the mean of the absolute discrepancy between these two quantities as a measure of the mean calibration error (MCE) (see e.g. \cite{ECP,jungo}).
For assessing if the uncertainty in the inferred parameters $\vec{\Omega}$ is consistent with the model's deviation from the PDE numerical solutions, we performed a Monte Carlo goodness-of-fit test (see e.g. \cite{weiss}).
This involves estimating an empirical p-value defined as the proportion of posterior samples whose deviation from the posterior mean exceeds that of the model curve.

\subsection{1D Poisson equation with a Gaussian source}

We consider a 1D Poisson equation equipped with a Gaussian source as follows. 
\be
\label{Poisson}
\frac{d^2 u}{dx^2} + 
e^{-\frac{(x-x_0)^2}{2\sigma^2_f}} = 0,
\ee
where $x_0, \sigma^2_f$ are parameters to be inferred from the model. 
The synthetic dataset was based on an exact solution $u_e (x)$ defined on the finite segment 
$x \in [0,1]$, with $x_0 = 1/3, \sigma^2_f = 0.02$. 
Dirichlet boundary conditions $u=0$ are imposed at 
$x = 0,1$. A uniformly distributed random noise of the form $\sim \mathcal{U}(-0.5,0.5)$ was added to the exact solution within $(0,1)$. We chose $\lambda_N$ to be 0.1 of the maximum value of $u_e$.
The size of the training and testing data was taken to be 240 and 150 (randomly chosen) points respectively covering the entire finite domain.\footnote{About half of the training dataset was used as a validation set for choosing the optimal learning rate in the first phase of model training, after which the whole training dataset was used.}

\begin{figure}[htp]
	\begin{center}		
\includegraphics[width=0.95\textwidth]{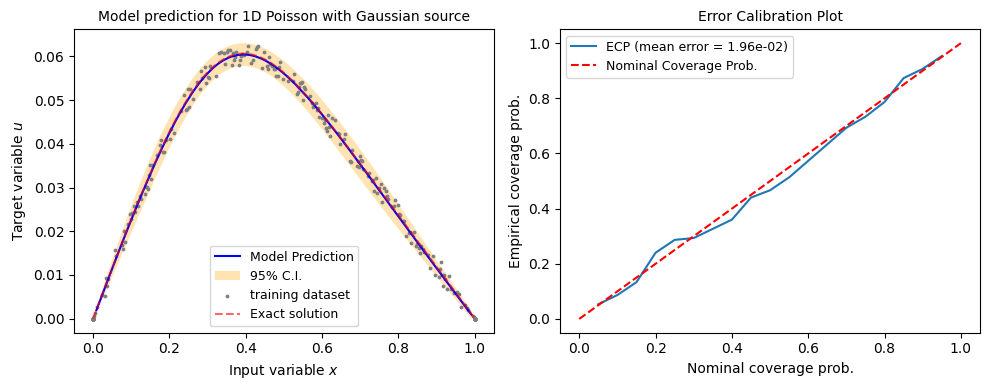}
	\end{center}
	\caption{\small The left figure shows the model prediction enclosed within a 95\% confidence band, together with the exact solution and training dataset. The right ECP vs NCP plot shows the model's uncertainty distribution to be well-calibrated. }
	\label{fig:poisson_final}
\end{figure}

For this problem, our base model was a fully-connected two-layer perceptron model with 16 neurons per layer.  
In the first training phase where the model was fitted purely to data, convergence was attained upon $2\times 10^5$ epochs at a learning rate of $10^{-4}$, with the Adam optimizer implemented via PyTorch. 
Using this initial model, we determined the prior distribution $\pi(\vec{\Omega})$ using eqns.~\eqref{prior_formula_1} and \eqref{stats_prior}, following our earlier discussion in Section 2.2.
Once the prior distribution $\pi (\vec{\Omega})$ was determined, we used it to determine the prior for $\sigma^2_R$ using eqn.15-18. 
The finite parameter domains were set to be $x_0 \in [0,1], \sigma^2_f \in [0.01, 0.06]$ discretized such that the minimal spacing is 1/50 of each interval. We then proceeded to the second phase of model training with a learning rate of $5\times 10^{-4}$.  
Upon completion of training, our model inferred the equation's parameters to be 
\be
\label{poisson_para}
x_0 = 0.334^{+0.013}_{-0.008},\,\,\,
\sigma^2_f = 0.0206^{+0.0027}_{-0.0024}
\ee
where the upper and lower figures refer to the 68\% confidence intervals. We note that they are consistent with the actual values of these parameters ($x_0 = 0.333, \sigma^2_f = 0.02$).
The p-value for the Monte Carlo goodness-of-fit test was 0.89, showing lack of evidence against the hypothesis that the model curve arises from the inferred posterior distribution characterized by \eqref{poisson_para}. 
The model's uncertainty distribution appeared to be well-calibrated, with the ECP close to the NCP for all confidence intervals. The mean calibration error was 0.02. Figure~\ref{fig:poisson_final} shows the model-predicted curve and the ECP vs NCP plot revealing a well-calibrated uncertainty distribution.

\subsection{1+1D Fisher-KPP equation}
\begin{figure}[htp]
	\begin{center}		   \includegraphics[width=\textwidth]{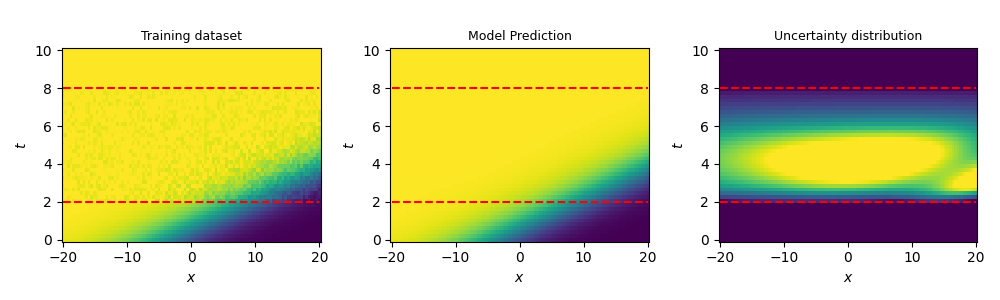}
	\end{center}
	\caption{\small  Heatmap representation of the model's prediction displayed next to the training dataset for comparison. Noise was added to the region bounded within the red dashed lines. The rightmost diagram shows the uncertainty distribution heatmap mirroring the noisy region in the training data. }    \label{fig:fisher_pred}
\end{figure}

As a second validation case study, we consider the Fisher-KPP equation with one spatial and a temporal dimension. 
\be
\frac{\partial u}{\partial t}
= D \frac{\partial^2 u}{\partial x^2} + r u \left( 
1 - \frac{u}{K}
\right),
\ee
where $r,D$ are parameters to be inferred. They physically represent the growth and diffusion rates respectively when this equation is used to model reaction-diffusion systems that can be applied to wave propagation and population growth dynamics, etc. We construct our synthetic dataset based on an exact traveling wave solution first presented by Ablowitz and Zeppetella in 
\cite{ablo}. 
\be
\label{fisher_exact}
u(x,t) = \left[
1 + \text{exp}
\left(
\sqrt{\frac{r}{6D}}
\left(
x - \frac{5\sqrt{rD}}{\sqrt{6}}t
\right)
\right)
\right]^{-2}.
\ee
We set $r = 1.6, D = 6.2$ for generating the training dataset. These parameters are the ones to be inferred from the model. Like in the Poisson equation case study, we added a uniformly distributed random noise $\sim \mathcal{U}(-0.5,0.5)$ to the exact solution \eqref{fisher_exact} defined on the 2D domain 
$x\in (-20,20), t\in (0,10)$ (see Fig.~\ref{fig:fisher_pred}).
The training and testing datasets were set to be $5\times 10^3$ and $4\times 10^3$ randomly distributed points across the domain. Like in the previous case study, 
we used a fully connected two-layer perceptron model with 16 neurons per layer. The training dynamics was similar to the previous case, with the model trained for $2\times 10^5$ epochs using only the data loss term and the prior distribution constructed from the HDR of the empirical distribution as defined in eqn.~\eqref{fund_prior}. The second phase of model training was implemented at a learning rate of $3\times 10^{-4}$, with convergence attained after $3\times 10^5$ epochs. At the end of training, our model inferred the equation's parameters to be
\be
\label{fisher_val}
r = 1.60^{0.05}_{-0.03},\,\,\,
D = 6.26^{1.22}_{-1.30}.
\ee
We note that the 68\% confidence intervals in \eqref{fisher_val} are consistent with the actual values $(r = 1.6, D = 6.2)$. The p-value for the goodness-of-fit test was 0.93, indicating lack of evidence against the posterior distribution of \eqref{fisher_val}. The model's uncertainty quantification appeared to be well-calibrated similar to the previous Poisson equation case study, with a mean calibration error of 0.024. 

\subsection{Comparison against other methods of uncertainty quantification frameworks in PINN}

We compare the performance of our model against Bayesian PINN and Deep Ensemble methods, the main metrics being the deviation of the empirical coverage probability from the nominal one, and whether the model could recover the actual PDE parameters within appropriate
confidence intervals. 

Deep Ensemble method was implemented with 100 
ordinary PINNs each formulated on the same 2-layer fully-connected perceptron trained with 
the ordinary mean-squared error loss function and the typical PINN residual loss function. The ensemble of models was defined by a random distribution in each model's initial weights. 
Bayesian PINNs (B-PINN) were constructed following the main principles presented in \cite{bpinn}. Perceptron weights and PDE parameters were treated as latent variables, and inferred using Markov-Chain-Monte-Carlo sampling (implemented in NumPyro and Wax).
Further we employed the No-U-Turn Sampler (NUTS), an adaptive variant of Hamiltonian Monte Carlo, which automatically adjusts the step size during warm-up to achieve efficient sampling and a desirable acceptance rate. For all models, training was performed on NVIDIA A100 GPU (40GB). In the aspect of computational cost, the average times taken for the completion of training for the Poisson, Fisher-KPP case studies were: (i) (160 min, 56 min) for E-PINN (ii) (70 min, 100 min) for B-PINN with $10^4$ posterior samples with 30 percent burn-in (iii) (4500 min, 4000 min) for Deep Ensemble. 

\begin{figure}[htp]
	\begin{center}		  \includegraphics[width=0.9\textwidth]{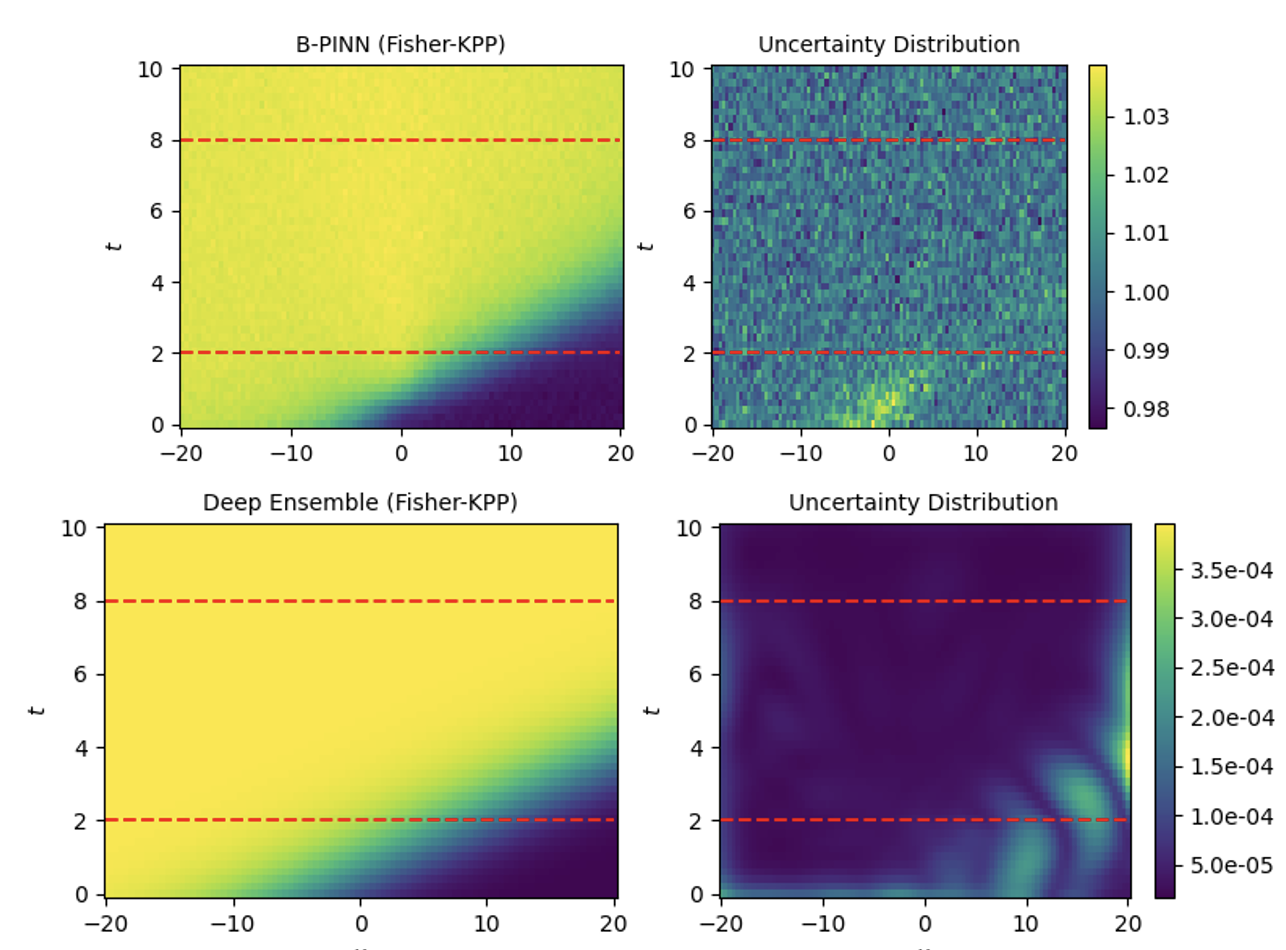}
	\end{center}
	\caption{\small Each row shows the model prediction and its associated uncertainty distributions for the B-PINN (top) and Deep Ensemble (bottom) methods. There is an absence of elevated uncertainty values within the red boundaries (where noise was inserted in the training data) unlike the case of E-PINN (Fig. 2), showing low correlation between uncertainty estimates and the added noise for these two methods. }
	\label{fig:U_BPINN_fisher}
\end{figure}

In terms of the degree of uncertainty calibration, we found E-PINN to perform significantly better than Deep Ensemble and B-PINN models for both case studies. While B-PINN tended to yield over-conservative uncertainty estimates, Deep Ensemble models generated excessively over-confident ones. The ECP vs NCP plots showed E-PINN's uncertainty estimates to be well-calibrated in contrast to the more obvious deviations from the NCP for the other two frameworks. The MCE values for E-PINN were much lower relative to B-PINN and Deep Ensemble. 
For the 2D Fisher-KPP case study in which heterogeneous noise was inserted into the training data, we found that the E-PINN uncertainty estimates yielded a strong correlation with the training data noise (Spearman's coefficient $r_s \sim 0.8, p<0.01$) which was absent in Deep Ensemble ($\rho_s \sim 0.01, p=0.2$) and B-PINN ($\rho_s \sim 0.09, p<.001$) as shown in the Fisher-KPP uncertainty plots (Fig.~\ref{fig:U_BPINN_fisher}).

In terms of the accuracy of the inferred parameters, all three frameworks recovered the unknown values within $\pm 1 \sigma$ of their respective predicted confidence intervals. For B-PINN, we found the inferred parameters to be the least accurate, whereas the Deep Ensemble yielded significantly smaller parameter uncertainties that proved to be problematic in the Fisher-KPP case, at least in the aspect of a low p-value for the goodness-of-fit test. These problems were consistent with an over- and under-confident nature of the uncertainty estimates associated with Deep Ensemble and B-PINN models respectively. 
In Table~\ref{tab:results} below, we tabulate all performance metrics for both case studies.

\begin{table}[htp]
  \caption{\small Summary of performance metrics for all three models.
  \emph{Abbrev.:} The first two rows are the learnable parameters for the 1D Poisson and 2D Fisher-KPP equations; MCE = mean calibration error for each of the two case studies; p-value: associated with the Monte Carlo goodness-of-fit test; $r_s$: Spearman's coefficient measuring correlation between uncertainty and noise.}
  \label{tab:results}
  \centering
  \begin{tabular}{r|ccc}
  \hline
    $\,$ & \textbf{E-PINN} & \textbf{B-PINN} & \textbf{Deep Ensemble} \\
    \hline \\
   ($x_0, \sigma^2_f$) & $0.334^{+0.013}_{-0.008},\,
0.0206^{+0.0027}_{-0.0024}$ & 
    $0.47^{+0.19}_{-0.19},\,
    0.029^{+0.014}_{-0.014}$
    & 
$0.334^{+0.0004}_{-0.0002},\,
 0.0198^{+0.00003}_{-0.00005}$
    \\ \\
    ($r,D$) & 
    $1.60^{+0.05}_{-0.03},\,
    6.26^{+1.2}_{-1.3}$
        & 
        $1.44^{+0.86}_{-0.85},\,
    5.26^{+2.2}_{-2.3}$
        & 
        $1.5968^{+0.0002}_{-0.0002},\,
    6.194^{+0.011}_{-0.010}$    
        \\ \\
    MCE & (0.02, 0.02) & (0.35, 0.49) & (0.44, 0.40) \\
    p-value  & (0.89, 0.93) & (0.02, 0.53) & (0.49, 0.003) \\
    $r_s$ coef. & 0.75, ($p<.01$) & -0.09 ($p<.01$) & 0.02 ($p=0.2$) \\
    \hline
  \end{tabular}
\end{table}

\section{Application of E-PINN to Bergman minimal model of glucose-insulin regulatory system}
\label{sec:Bergman}

In this Section, we present a brief case study of E-PINN modeling published clinical data \cite{Gaetano,Kartono,Ward} related to the intravenous glucose tolerance test (IVGTT) -- a procedure that has been widely used in diabetes-related physiological research \cite{bergman3}. Within the literature related to mathematical models of glucose-insulin interactions, many works have been constructed based on a model due to Bergman et al. \cite{bergman2,bergman1} which is the following set of coupled first-order differential equations representing glucose-insulin dynamics. 
\bea
\label{bergman1}
\frac{dG}{dt} &=& -p_1
(G - G_b)- X(t) G(t), \\
\label{bergman2}
\frac{dX}{dt} &=& -p_2 X(t) + p_3 (I(t) - I_b),
\eea
where $G$ represents glucose level, $I$ represents insulin concentration and $X$ a more abstract quantity representing glucose uptake activity of insulin-excitable tissue (e.g. liver, adipose, etc.). $G_b, I_b$ are the basal concentrations of glucose and insulin representing baseline values when the effect of the bolus injection of glucose in IVGTT disappears. The parameters $p_1, p_2, p_3$ represent the glucose decay rate, spontaneous decay rate of remote insulin and the insulin-dependent increase in tissue glucose uptake ability respectively. These are the parameters we leverage E-PINN to infer from measurements of glucose and insulin evolutions in IVGTT samples. In particular, $p_1$ is known as an index measuring `glucose effectiveness' (often denoted as $S_G$), while $S_I \equiv p_3/p_2$ is known as the insulin sensitivity index that measures the effectiveness at which glucose decreases upon release of insulin. In a IVGTT test, glucose and insulin samples are periodically collected over a period of a few hours. Taking the insulin to be an externally known function, the equations \eqref{bergman1}, \eqref{bergman2} thus constitute a coupled set of first-order ODEs in the variables $G$ and $X$ which is often alluded to as the \emph{Bergman minimal model} \cite{bergman2}.

\begin{figure}[htp]
\begin{center}		
\includegraphics[width=1.05\textwidth]{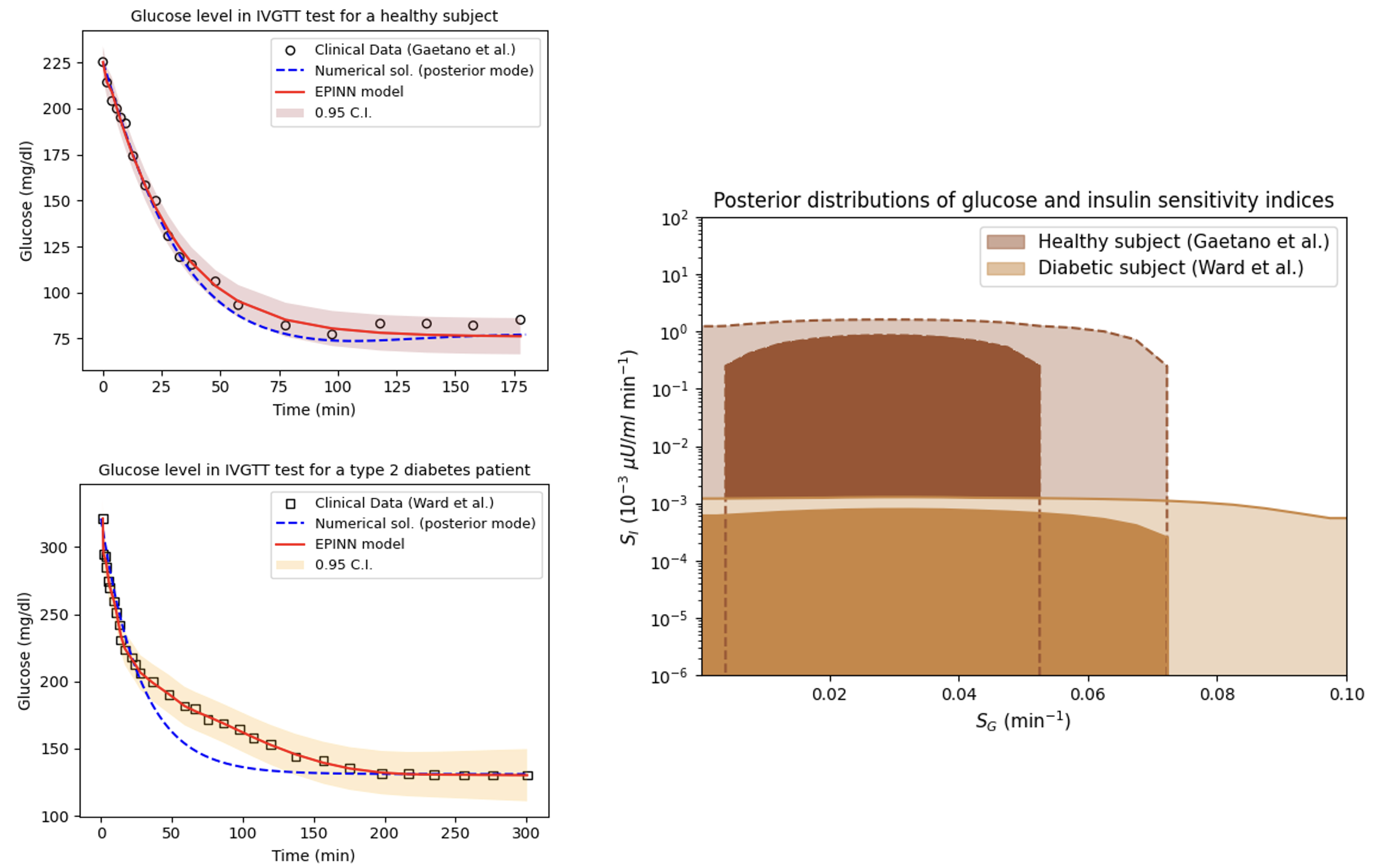}
\end{center}
\caption{\small The left diagrams show E-PINN's prediction curves for glucose evolution in healthy and diabetic subjects. Blue dashed curves are the numerical solutions to Bergman's ODE with parameters $p_1, p_2, p_3$ being the posterior modes. Deviations between blue and red curves in the middle regions suggest that PDE description may be relatively less valid during those durations. The right diagram superposes the learned posterior distributions of the indices $S_G, S_I$, revealing the latter as a more discriminatory index for identifying diabetic individuals. Log scale for $S_I$ was adopted to enable clearer visualization of the distributions for healthy (dark brown) and diabetic (light brown) patients.}
\label{fig:bergman_curves}
\end{figure}

We apply E-PINN to the modeling of \eqref{bergman1} and \eqref{bergman2}, using the datasets gathered and analyzed in \cite{Gaetano,Kartono,Ward}. In particular, we focused on modeling two datasets pertaining to the measured glucose and insulin levels corresponding to healthy and type 2 diabetic subjects. E-PINN enables probabilistic modeling of the datasets and directly yields the posterior distribution of the unknown parameters. 
Unlike the controlled case studies, we do not have empirical data points for the variable $X(t)$ and thus, our base neural network was designed to yield five outputs -- four corresponding to $G(t)$ and its uncertainty, and another corresponding to $X(t)$ so as to avoid overfitting on $X(t)$. We define the PDE residual as the sum of the residual arising from \eqref{bergman1} and \eqref{bergman2} separately. Estimating $I_b, G_b$ from the datasets, we are left with only $p_1, p_2, p_3$ as the unknown parameters to be inferred. Fig. \ref{fig:bergman_curves} depicts our model's prediction curves and the learned posterior distributions of two clinically useful indices: the glucose effectiveness index $S_G$ and insulin sensitivity index $S_I$, the latter appearing to be the more discriminatory one, being about 3 orders of magnitude larger for healthy individuals.

\section{Conclusion}
\label{sec:conclusion}
We have constructed a class of PINN that is equipped with an uncertainty quantification framework for its target and PDE parameter inferences. It leverages the loss function of EDL for estimating uncertainty of outputs and infers unknown PDE parameters through learning a posterior distribution. Two key features distinguish our model from the usual PINN: (i)the loss weight of the PDE residual term evolves throughout training, (ii)a prior distribution based on a purely data-fitted model regularizes learning of the unknown parameters. Our methodology is broadly applicable to systems of differential equations, irrespective of the choice of differential operators or dimensionality.
Validating our model on two controlled case studies involving the 1D Poisson equation with a Gaussian source and the 2D Fisher-KPP equation, we found that E-PINN generated empirical coverage probabilities that were calibrated significantly better than the B-PINN and Deep Ensemble methods. It inferred unknown parameters accurately within the bounds of confidence intervals consistent with the Monte-Carlo goodness-of-fit test. In the Fisher-KPP case, E-PINN's uncertainty distribution correlated well with a temporally varying noise, unlike the other two frameworks. As an illustrative real-world application, we demonstrated how E-PINN can be applied to model clinical IVGTT datasets presumed to follow Bergman minimal model of glucose-insulin dynamics, with results indicating that the insulin sensitivity index $S_I$ discriminates effectively between healthy and diabetic patients. We look forward to extensive applications of E-PINN to solving inverse problems across scientific disciplines.

\bibliographystyle{splncs04}
\bibliography{TAAI2025}

\end{document}